\documentclass{article} 
\usepackage{style/arxiv}

\usepackage{url}
\usepackage{import}
\usepackage{multirow}
\usepackage{subcaption}
\usepackage{graphicx}
\usepackage{amsmath}

\usepackage[pagebackref=true,breaklinks=true,letterpaper=true,colorlinks,bookmarks=false]{hyperref}

\begin{document}

\title{Ocular Authentication: Fusion of Gaze and Periocular Modalities}

\author{
    Dillon~Lohr\textsuperscript{1}, 
    Michael J. Proulx\textsuperscript{1}, 
    Mehedi Hasan Raju\textsuperscript{2},  
    Oleg V. Komogortsev\textsuperscript{1,2}\\
    \textsuperscript{1}Meta Reality Labs Research, Redmond, WA, USA, 
    \textsuperscript{2}Texas State University, San Marcos, Texas, USA \\
    {\tt\small dlohr@meta.com, michaelproulx@meta.com, m.raju@txstate.edu, ok@txstate.edu}
}
\date{}

\maketitle

\begin{abstract}

This paper investigates the feasibility of fusing two eye-centric authentication modalities—eye movements and periocular images—within a calibration-free authentication system. 
While each modality has independently shown promise for user authentication, their combination within a unified gaze-estimation pipeline has not been thoroughly explored at scale. 
In this report, we propose a multimodal authentication system and evaluate it using a large-scale in-house dataset comprising 9202 subjects with an eye tracking (ET) signal quality equivalent to a consumer-facing virtual reality (VR) device.
Our results show that the multimodal approach consistently outperforms both unimodal systems across all scenarios, surpassing the FIDO benchmark.
The integration of a state-of-the-art machine learning architecture contributed significantly to the overall authentication performance at scale, driven by the model's ability to capture authentication representations and the complementary discriminative characteristics of the fused modalities.

\end{abstract}


\section{Introduction}

The proliferation of extended reality (XR) device usage has introduced both new opportunities and challenges for authentication security.
As XR applications expand across various domains, ensuring secure and continuous user authentication becomes increasingly critical \cite{lohr2024baseline}. 
This demand closely aligns with the foundational principles of authentication systems.
Biometric systems fundamentally rely on the distinctiveness and repeatability of an individual's physiological (e.g., fingerprint, iris, facial structure) and behavioral (e.g., gait, voice, eye movement) traits \cite{jain2007handbook, raju2024, kumari2022periocular}.

Many XR headsets are already equipped with integrated ET technologies designed to optimize rendering \cite{foveated_rendering,foveated_rendering2} and enable gaze-based interaction \cite{piumsomboon2017exploring, dondi2023gaze, raju2025_gaze_interaction}.
This native support for ET system opens the door to unimodal (single trait) authentication solutions, such as eye movement-based authentication (EMA) and periocular-image-based authentication (PIA). 

EMA uses the person-specific information from the ET signal to authenticate the individual based on eye movements \cite{eky,ekyt}, whereas PIA extracts discriminative features from the images of the periocular region (including the iris, eyelid, eyelashes, eyebrow, tear duct, eye shape, skin texture, and many more) \cite{kumari2022periocular}.
Each of these modalities shows promise in unimodal authentication system \cite{lohr2023demonstrating, raju2024evaluating, lohr2024baseline, park2009periocular,alonso2024periocular, bhamare2025muswin}, but there is still the possibility to enhance the authentication and spoofing resistance performance by combining them. 
Notably, multimodal authentication systems have gained attention for improving authentication accuracy over unimodal systems \cite{pahuja2024multimodal}. 
In the previous research \cite{komogortsev2015,raju2022iris}, it was demonstrated that EMA is effective in detecting print attacks and mechanical replicas, but investigating this aspect of performance is beyond the scope of this work.

Existence of eye tracking in XR devices provides an opportunity to employ eye images captured by a gaze estimation pipeline and the gaze itself for accurate and spoof-resistant authentication.  
This is particularly advantageous because traditional multimodal systems often face challenges such as increased hardware complexity and power consumption, reduced usability, and higher deployment costs due to the reliance on multiple sensors and extensive data acquisition requirements \cite{oleg2012}.

Although prior work has evaluated unimodal systems using either eye movements or periocular images—and some have explored multimodal authentication systems involving other discriminative traits—there is limited research on combining EMA and PIA within a unified framework using data readily available from standard XR gaze estimation pipelines.
In this work, we address this gap and report the feasibility of multimodal user authentication combining periocular image and gaze (eye-movements) captured by the same gaze estimation pipeline, which is specifically optimized for eye tracking performance. 
The device used to collect the dataset employed in this work offered ET signal quality comparable to that of the Meta Quest Pro \cite{wei2023preliminary, aziz2024evaluation}, making it a suitable representative of consumer-facing VR platforms.

To the best of our knowledge, this study is the first to explore multimodal authentication of gaze and periocular modalities on a dataset of almost ten-thousand people, with major outcomes indicating strong benefit of their fusion.

The following research questions guide our investigation:

\paragraph{(RQ1) Does multimodal fusion of eye movement and periocular image improve authentication accuracy over single modality baselines?}  
We compare the authentication performance of EMA and PIA authentication systems independently and assess the gains achieved through their fusion.
\paragraph{(RQ2) Which fusion strategy contributes more effectively to performance improvement?}  
We explore both score-level and embedding-level fusion strategies to determine which approach yields superior authentication accuracy in the multimodal setup.
\paragraph{(RQ3) How does embedding aggregation influence authentication accuracy?} 
We conduct controlled experiments to analyze the effects of varying the duration of gaze input and the number of periocular image samples on resulting performance.

\section{Prior Work}

\subsection{EMA \& PIA}

Kasprowski and Ober’s \cite{Kasprowski2004} pioneer work established eye movements as a viable biometric modality for user authentication, laying the foundation for a growing body of research in this domain. 
Since then, numerous studies have contributed to the advancement of eye movement-based authentication systems, exploring various methodologies and applications \cite{
bednarik2005eye, holland2011, komogortsev2012biometric, holland2013, komogortsev2014biometrics, zhang2015survey, nigam2015ocular, zhang2016biometrics, galdi2016eye, friedman2017method, RIGAS2017129, kasprowski2018fusion, li2018biometric, brasil2020eye, katsini2020role, liao2022exploring, raju2024, grandhi2025evaluating, EmMixformer2025}.

One of the primary reasons for the continued interest in eye movement data is its inherent resistance to imitation and spoofing. 
Because these movements are governed by both cognitive and physiological processes, they are difficult to replicate with precision \cite{deepeyedentificationlive, ekyt, lohr2020metric, Lohr2020, deepeyedentification, rigas2015, raju2022iris, komogortsev2015}. 
As the field evolved, machine learning techniques—especially deep learning—have played an increasingly central role in gaze-based authentication systems. 
Two dominant methodological directions have emerged: systems that rely on pre-extracted, handcrafted features \cite{lohr2020metric, george2016score} and end-to-end models that learn discriminative representations directly from raw ET data \cite{Jia2018, eky, deepeyedentification, deepeyedentificationlive, ekyt}.

More recently, EMA has been evaluated in XR, where users interact in naturalistic and unconstrained settings. 
This line of work has shown that even lower-end consumer-facing eye trackers embedded in head-mounted displays can achieve competitive authentication performance \cite{lohr2018implementation, lohr2020eye, lohr2023demonstrating, raju2024evaluating, lohr2024baseline}. 
Notably, state-of-the-art systems that employ end-to-end pipelines and large-scale gaze datasets have achieved remarkably low equal error rates—reaching as low as 0.04\% in some cases \cite{ekyt, lohr2024baseline}. 
These findings demonstrate that eye movements have matured into a meaningful-performance, standalone capable behavioral biometric modality.

Parallel to the expansion of EMA research, the periocular region—encompassing the area surrounding the eye—has gained traction as a valuable biometric trait, particularly in situations where traditional iris recognition is compromised \cite{kumari2022periocular}. 
First introduced as a distinct modality in 2009 \cite{park2009periocular}, periocular biometrics offer resilience under challenging conditions such as occlusion, non-uniform lighting, pose variation, and limited subject cooperation \cite{woodard2010periocular}. 
Since then, several studies have explored the discriminative potential of the periocular region, contributing to its advancement as a trait to be considered in a bio-authentication system \cite{hollingsworth2010identifying, woodard2010periocular, park2010periocular, hollingsworth2011human, kumari2022periocular, ambika2016periocular, zanlorensi2022new, alonso2024periocular, bhamare2025muswin}.

Recent innovations have integrated periocular features with modern deep learning techniques. For instance, EyePAD++ \cite{eyepad++} introduced a knowledge distillation-based framework that simultaneously addresses eye-based authentication and presentation attack detection. This unified approach achieved verification performance comparable to state-of-the-art methods while also enhancing system robustness against spoofing attempts.

\subsection{Toward Multimodal Authentication System}

\begin{figure*}[htbp]
\centering
\includegraphics[width=\textwidth]{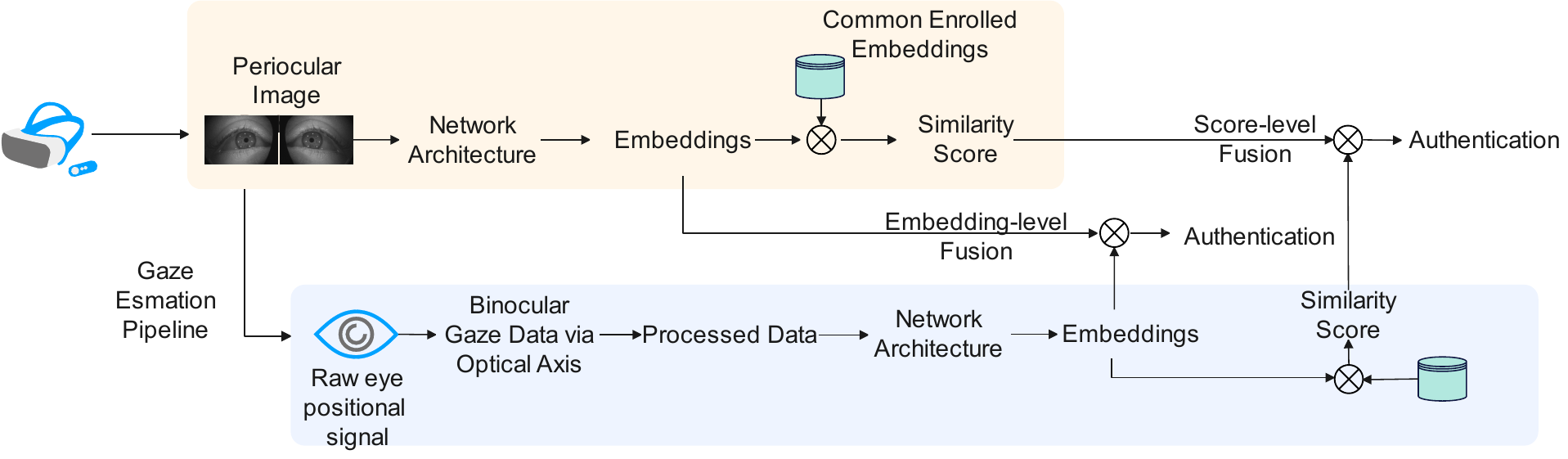}
\caption{Ocular Multimodal Authentication System}
\label{fig:multimodal_approach}
\end{figure*}

While EMA and PIA have independently demonstrated significant promise, the integration of these two modalities into a unified multimodal framework remains unexplored though they have combined with other traits.
In the case of EMA, researchers have explored fusion with other modalities such as iris patterns \cite{oleg2012}, touchscreen interaction behavior on mobile devices \cite{et+touch, et+touch2}, PIN entry dynamics \cite{et+pin}, and EEG signals \cite{et+eeg}. 
Some of the most recent efforts have even examined the potential of combining gaze data with brainwave activity within XR environments \cite{et+bw}, highlighting the modality's adaptability across diverse platforms.
Similarly, periocular features have been integrated with scleral characteristics \cite{p+sclera}, facial and ear images \cite{p+face, p+face2, p+face3, p+face_ear}, and eye blink signals \cite{p+eye_blink}, showcasing the usefulness of the periocular region in multimodal contexts.

Despite this growing interest in multimodal systems, there is currently no published research that explicitly investigates the fusion of EMA and PIA. This absence is notable given that both modalities originate from the same ocular region and offer complementary strengths—one being behavioral with information coming from dynamic aspects of eye motility in the time domain, while the other is represented by a static signal coming from spatial image domain. 
ET gaze estimation pipeline provides a perfect opportunity to explore the fusion of both without the necessity of any additional hardware.
Combining them could improve authentication accuracy, because dynamics of eye movements and periocular features are uncorrelated to each other.

Motivated by this gap in the literature, our work aims to be the first to explore this fusion within a unified authentication framework. 
By examining the interplay between these eye-centric modalities, we seek to establish whether their integration can unlock new levels of authentication performance.


\section{Multimodal Authentication Approach}

In this research, we proposed a multimodal authentication system that integrates periocular images and eye movement behavior for enhanced user authentication.

Fig.\ref{fig:multimodal_approach} illustrates a general overview of our proposed multimodal approach.
The upper stream processes periocular image captured from both eyes using a dedicated network architecture that extracts feature embeddings and computes similarity scores against known user templates. 
Simultaneously, the lower stream processes raw binocular gaze data, only using the optical axis. 
These signals are preprocessed and passed through a densenet-based network architecture to generate discriminative embeddings, which are compared with enrolled gaze signatures. 
The outputs from both modalities are fused at both the score- and embedding-level to produce a final authentication decision.

\subsection{Dataset}

We employed GazePro, an internal dataset of gaze signals \cite{lohr2024baseline}, collected at 72~Hz with ET signal quality similar to Meta Quest Pro \cite{wei2023preliminary,aziz2024evaluation}.
The appearance-based gaze estimation approach for the device employed to capture the data was similar to the approach of the Project Aria Eye Tracking by Facebook Research\footnote{\href{https://github.com/facebookresearch/projectaria_eyetracking}{https://github.com/facebookresearch/projectaria\_eyetracking}}.

GazePro includes 9,202 participants: 3,673 self-reported as male, 5,376 as female, and 153 as neither. 
Ages ranged from 13 to 88 years (median = 33, IQR = 18). 
All had normal or corrected vision, with 3223 wearing glasses or contact lenses and 5979 wearing no corrective lenses.
A total of 2,032 participants were reported to be wearing some form of eye makeup, including eyelash extensions, heavy eyeshadow, and/or mascara. Regarding eye color distribution, 5,614 participants had brown eyes, followed by 1,759 with blue, 1,113 with hazel, 701 with green, 13 with amber, and 2 with violet eyes.

Participants did not utilize a chin rest during data collection. 
The experimental protocol comprised a series of tasks, including random saccades, smooth pursuit, and vestibulo-ocular reflex tasks—each involving a single focal target designed to promote sustained visual attention. 
The total duration of the session was approximately 20 minutes. 
Several tasks were modeled after those described by Lohr et al. \cite{gazebasevr}.

Participants were occasionally instructed to perform various facial expressions—such as winking, blinking, squinting, laughing, expressing anger, and displaying sadness—to introduce natural facial dynamics into the recordings. 
Additionally, at designated intervals, subjects were asked to adjust the position of the VR headset, resulting in slight shifts in device placement.
Throughout the session, an experiment facilitator was present to ensure participants adhered closely to task instructions and to maintain high data quality.

\begin{figure}[htbp]
\centering
\includegraphics[width=0.8\textwidth]{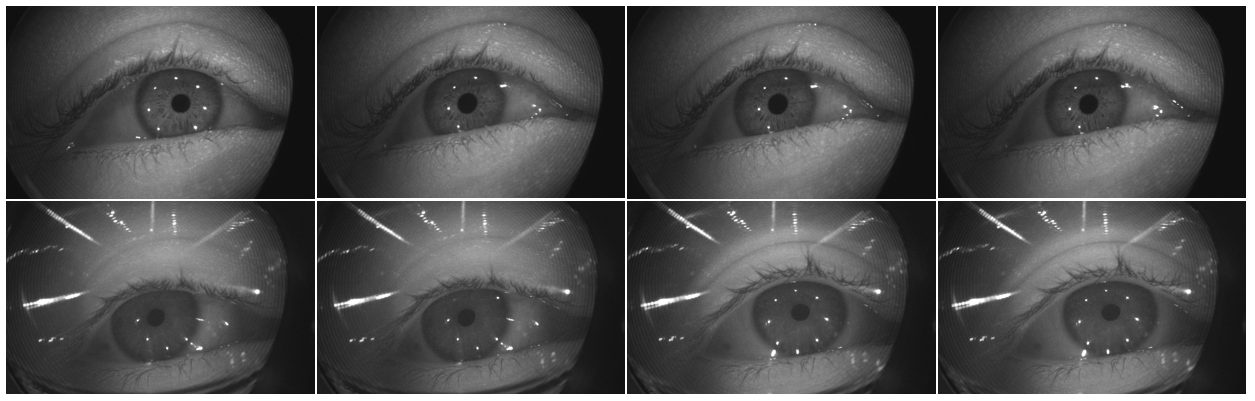}
\caption{Exemplar periocular images taken from OpenEDS dataset: Normal vision (Top-row), Corrected vision (Bottom-row).}
\label{fig:periocular_samples}
\end{figure}

Notably, GazePro features a significantly larger participant pool than previous datasets in this domain \cite{gazebase,gazebasevr,makowski2020biometric}, enabling broader evaluation.
The periocular images employed for this study are similar to that of the OpenEDS dataset \cite{openeds2020}.
Fig. \ref{fig:periocular_samples} shows examples of periocular images.

\subsection{Eye-movement driven Authentication (EMA)}

\paragraph{Network Architecture---} We utilized Eye Know You Too (EKYT) \cite{ekyt}, a state-of-the-art user authentication model based on DenseNet architecture \cite{densenet}. 
EKYT has demonstrated superior performance in EMA authentication, particularly with high-volume of data collected at 72 Hz, achieving an equal error rate (EER) of 0.04\% using both visual and optical axes estimation.
The architecture comprises eight convolutional layers, where feature maps generated by each layer are concatenated with those from all preceding layers before being forwarded to the next convolutional layer. 
The final concatenated feature maps are subsequently flattened, subjected to global average pooling, and processed through a fully connected layer to produce 128-dimensional embeddings.
For a comprehensive description of the network architecture, refer to \cite{ekyt}.

In gaze-study, we closely adhered to the training methodology used for \cite{lohr2024baseline}, with optical-axis estimation only.

\paragraph{Data Processing---} Before being fed into the network, all dataset recordings underwent a series of preprocessing steps.
We derived two velocity channels (horizontal and vertical) from the raw signal using a Savitzky-Golay filter \cite{savitzkyGolayM} with a window size = 7 and order = 2 \cite{friedman2017method}. 
The data were then segmented into non-overlapping 5-second windows (360 samples) using a rolling window approach. 
For further evaluation, multiple of these 5-second segments were combined to make 10~seconds, and 20~seconds long.
To mitigate high-frequency noise, velocities were clamped between ±1000°/s. 
Finally, clamped velocities are standardized to have zero mean and unit variance using the transformation $(x - \mu)/\sigma$, where $\mu$ and $\sigma$ represent the mean and standard deviation (SD), respectively, computed from the training set.
Finally, NaN values were replaced with 0 as part of the data-handling process.
Further details on preprocessing can be found in \cite{ekyt}.

\paragraph{Model Training---}
The model is trained using the multi-similarity (MS) loss function~\cite{Wang2019}, which is defined as:

\begin{align}
\mathcal{L}_{\text{MS}} = \frac{1}{m} \sum_{i=1}^{m} \Bigg[ 
    &\frac{1}{\alpha} \log \left( 1 + \sum_{k \in \mathcal{P}_i} \exp \left[ -\alpha (S_{ik} - \lambda) \right] \right) \nonumber \\
  +\; &\frac{1}{\beta} \log \left( 1 + \sum_{k \in \mathcal{N}_i} \exp \left[ \beta (S_{ik} - \lambda) \right] \right) 
\Bigg],
\end{align}

where $m = 256$ is the minibatch size; $\alpha = 2.0$, $\beta = 50.0$, and $\lambda = 0.5$ are hyperparameters; $\mathcal{P}_i$ and $\mathcal{N}_i$ denote the sets of mined positive and negative pairs for the anchor sample $x_i$; and $S_{ik}$ represents the cosine similarity between samples $x_i$ and $x_k$. This formulation of MS loss implicitly incorporates an online pair mining strategy with an additional hyperparameter $\epsilon = 0.1$. Further details on MS loss can be found in~\cite{Wang2019}.

Each minibatch is constructed by randomly selecting 16 unique users from the training set and sampling 16 distinct examples per user, yielding $16 \times 16 = 256$ samples per minibatch. 
One training epoch consists of as many such minibatches as needed to match the total number of unique samples in the training set; however, not every sample is guaranteed to appear in each epoch. 

Training is conducted over 100 epochs using the Adam optimizer~\cite{adam} with a one-cycle cosine annealing learning rate scheduler~\cite{smith2019super}. 
The learning rate starts at $10^{-4}$, increases to a peak of $10^{-2}$ over the first 30 epochs, and then decays to a minimum of $10^{-7}$ across the remaining 70 epochs.





\subsection{Periocular Image driven Authentication (PIA)}

We have partially followed the approach proposed in \cite{eyepad++} for user authentication using periocular images.
In this approach, no image preprocessing steps were considered, thus excluding the step of detecting or segmenting the iris region. 
The entire periocular image is considered as input to the network.
Our network architecture is similar to that proposed in \cite{eyepad++} for periocular image-based authentication purposes.
This architecture used Densenet121 as a backbone \cite{densenet}.
The developed network is trained on periocular images from the dataset to support a PIA framework. 
Our training procedure is similar to  \cite{eyepad++}.
For details about network architecture and training procedure, please follow \cite{eyepad++}.

\subsection{Performance Evaluation}

For EMA at evaluation time, we select one set of ET recordings—at most one per participant—for enrollment and another, disjoint set—again, at most one per participant—for verification. 
Each ET recording undergoes the same preprocessing steps as during training: recordings are segmented into non-overlapping 5-second chunks, velocities are computed using a Savitzky-Golay differentiation filter~\cite{savitzkyGolayM}, and clamped to the range of $\pm1000^\circ$/s. 
The resulting velocities are then standardized using the mean and standard deviation computed over the training set.

The gaze estimation pipeline used in GazePro provides valid gaze outputs for all time points, including during eye blinks. 
Therefore, we omit the minimum gaze validity criteria used in prior work~\cite{ekyt}, as it was unnecessary in this context.
The first $n$ valid 5-second chunks from each ET recording are passed independently through the model to generate $n$ embeddings. 
These embeddings are then averaged to produce a single centroid embedding per recording. 
This process is repeated for all enrollment and verification recordings, resulting in two sets of centroid embeddings.

In PIA, a known user template is created initially for a known user by processing multiple left and right eye images through a CNN to extract deep feature representations. 
These feature vectors are then averaged separately for each eye to obtain representative feature embeddings.
During the verification phase, a pair of eye images—left and right—are similarly processed through the CNN to generate  256-dimension embedding.

To simulate verification, we compare 128-dimension gaze embeddings and 256-dimension image embeddings from the verification set against that from the enrollment set using cosine similarity. A verification attempt is considered \textit{genuine} if both embeddings originate from the same individual and \textit{impostor} otherwise.


\subsection{Performance Metrics}

The resulting similarity scores and associated labels are used to compute a Receiver Operating Characteristic (ROC) curve. 
From the ROC curve, several performance metrics are derived. 
We evaluated model performance using two key metrics: \textit{Equal Error Rate} (EER) \cite{rigas2015,evgeniy_rank} and False Rejection Rate (FRR) at a fixed False Acceptance Rate (FAR) \cite{conrad2015cissp, Quality2012}.  

EER represents the point on the ROC curve where FRR and FAR are equal. A lower EER indicates better user authentication performance. 
If such a point does not exist exactly, it is estimated using linear interpolation. To compute EER, we require separate enrollment and authentication datasets. 

FRR at a specified FAR threshold, denoted as $\text{FRR}_{X\%}$, where $X\%$ is the FAR expressed as a percentage. For example, $\text{FRR}_{0.002\%} = 1\%$ indicates that the system falsely rejects genuine users 1\% of the time when the FAR is constrained to 0.002\% (1-in-50,000).

In summary, EER provides a single performance measure, while FRR assesses real-world feasibility against standards such as the FIDO \cite{FIDO2020}.

\subsection{Fusion Approach}

We employed two broad fusion approaches in our study: (1) Score-level fusion (2) Embedding-level fusion.
Within each category, we implemented multiple techniques to compare.

\subsubsection{Score-Level Fusion (SF)}

\paragraph{Weighted Sum of Similarity Scores (SF1)---}

This approach uses a weighted sum fusion method inspired by \cite{ross2003information} to combine similarity scores from ET data and periocular images. 
The total contribution of both modalities is constrained to 1, ensuring a balanced evaluation. The weighted sum of similarity scores is computed as:

\begin{equation}
    S_f = w_p S_p + w_g S_g
\end{equation}

where \( S_p, S_g \) are the similarity scores from periocular and gaze respectively, \( w_p, w_g \) are the corresponding weights, ensuring \( w_p + w_g = 1 \), and \( S_f \) is the fused similarity score.

The weight assignments are systematically varied as:
\[
w_g = \{0.0, 0.1, 0.2, \dots, 0.9, 1.0\}, \quad w_p = 1 - w_g
\]

This allows testing different contribution levels from each modality to identify the optimal weight combination.

Each modality is also evaluated independently:
\begin{itemize}
    \item \textbf{EMA}: \( S_f = S_g \) (Weight: \( w_p = 0, w_g = 1 \)).
    \item \textbf{PIA}: \( S_f = S_p \) (Weight: \( w_p = 1, w_g = 0 \)).
\end{itemize}
This establishes a baseline for individual modality performance.

\paragraph{Rank-Based Fusion Approach (SF2)---}

For SF2, we followed the method proposed in Rigas et al. \cite{evgeniy_rank}.

This method assigns weights based on the overall ranking performance of an algorithm. 
It evaluates how frequently an algorithm ranks the correct identity higher in the similarity score ranking, ensuring that algorithms consistently placing correct matches at better ranks receive higher weights.

Given a ranked list \( R \) of similarity scores for a probe sample, let \( r_i \) be the rank position of the first correct match in the list. The weight for that match is computed as:    
$w(i) = 1 - \frac{r_i - 1}{|R|}$
where \( r_i = 1 \) gives maximum weight since the match is ranked first and as \( r_i \) increases, the weight decreases proportionally.
The final weight for matcher \( m \) is obtained by averaging the assigned weights over \( K \) probe items:
$w'_m = \frac{\sum_{i=1}^{K} w(i)}{K}$

Weights are then normalized so that their sum equals 1 using 
$w^{Rank}_m = \frac{w'_m}{\sum_{m=1}^{M} w'_m}$

The weights calculation is further transformed aiming at optimization.

$w^{Rank-opt}_m = \frac{w_m - min_m(W)}{max_{m}(W) - min_{m}(w)} - (1- w_{opt}) + w_{opt} $ 
Here, \(w_m\) is the weight of a specific matcher, W is the set of weights from all the matchers, and \(w_{opt}\) is the optimization parameter \cite{evgeniy_rank}. 

A special case of this method is Rank-1 based, which focuses on prioritizing algorithms that correctly match the identity at the first rank. 
It assigns full weight to algorithms that place the correct identity at Rank-1, and lower weights to others.

For a probe sample where the correct match appears at Rank-1, the assigned weight is:

$w(i) =
\begin{cases}
  1, & \text{if } r_i = 1 \\
  0, & \text{otherwise}
\end{cases}$

In our study we have two versions of SF2: 1) Rank-Opt 2) Rank-1-Opt.

\subsubsection{Embedding-Level Fusion (EF)}

\paragraph{Fused Embeddings (EF1)---} 

We implemented a multilayer perceptron (MLP) with no hidden layers in this approach, i.e., linear transformation. 
The model's input consists of embeddings from both modalities, resulting in an input size of \( 128 + 256 = 384 \). 
The model outputs fused embeddings with a dimensionality of either  \( 32, 64, 128,\) or \(256\).

The model was trained using Multi-Similarity Loss \cite{Wang2019} (without a miner) and optimized with Adam \cite{adam}, using a fixed learning rate of \( 3 \times 10^{-4} \). 
Training was conducted for up to 1000 epochs, utilizing all available data in a single batch. 
Validation results were computed every 100 epochs, and the best-performing model—based on the FRR at $1/50,000$ FAR (denoted by FRR$_{0.002\%}$)—was selected for reporting.
The training was performed on half of the test subjects (\( N = 953 \)) and validated on the remaining half (\( N = 953 \)). 
Results are reported for the validation set, differing from previous results, which were computed across all available test subjects.

Prior to concatenation and input into the model, gaze and periocular embeddings were L2-normalized. 
A fixed random seed of 42 was used to ensure reproducibility.
The similarity score was calculated on the fused embeddings.

\paragraph{Concatenation of Embeddings (EF2)--- }

Embeddings from the gaze-based authentication pipeline have been concatenated to 
that of the periocular images-based pipeline.
Embeddings are L-2 normalized before concatenation.
The similarity score was calculated on the concatenated embeddings.

\subsection{Experimental Set-up}

In this study, we designed two distinct experimental scenarios to evaluate the effectiveness of our user authentication:
\begin{enumerate}
    \item \textbf{Experiment 1: Single Periocular Image} — In this setup, we used a single periocular image per eye for each authentication attempt. 
    The images were carefully selected to ensure that the eye was fully open and unobstructed. This scenario reflects a minimal input setting, focusing on baseline performance with limited periocular data.

    \item \textbf{Experiment 2: Averaged Periocular Image Embedding} — This scenario was designed to explore the impact of embedding stability and robustness using multiple periocular images as inputs. 
    For all cases involving the periocular modality, we used five periocular images per eye. 
    Embeddings are aggregated by taking the mean of each feature (same as for gaze embeddings).
    We ensured that the pupil center was not occluded in any of the selected images, and the eye remained open in all cases. 
    The gaze-only modality was kept consistent with Experiment 1, serving as a common reference across both experiments.
\end{enumerate}

\begin{table*}[htbp]
\centering

\caption{Performance metrics across different experiment and fusion strategies. 
Lower metric values indicate better performance. 
Identical values across all rows within a block are marked with $^\dag$.  
``Same as above" refers to values carried over from Experiment-1.
The best-performing values are highlighted in bold.
}

\subfloat[EER (\%)]{
\resizebox{0.7\textwidth}{!}{%
\centering
\begin{tabular}{cccccccc}
\hline
Experiment & Gaze Length & EMA & PIA & SF1 & \multicolumn{2}{c}{SF2} & EF1 \\
\cline{6-7}
 & & & & & Rank-Opt & Rank-1-Opt & \\
\hline
\multirow{4}{*}{1} 
~& 5 & 20.4 & {\multirow{3}{*}{0.14$^\dag$}}  & 0.13 & 0.14 & 0.14 & 0.06\\
~& 10 & 12.6 &~& 0.12 & 0.13 & 0.13 & 0.06 \\
~& 15 & 8.03 &~& 0.12 & 0.12 & 0.12 & 0.06 \\
~& 20 & 5.75 &~& 0.1 & 0.1 & 0.1 & 0.05 \\
\hline
\multirow{4}{*}{2} 
~& {\multirow{3}{*}{Same as above}} & {\multirow{3}{*}{Same as above}} & {\multirow{3}{*}{0.12$^\dag$}} & .12 & 0.12 & 0.12 & {\multirow{3}{*}{\textbf{0.05}$^\dag$}}\\
~& & &~& 0.11 & 0.11 & 0.11 & \\
~& & &~& 0.1 & 0.1 & 0.1 & \\
~& & &~& 0.06 & 0.07 & 0.07 & \\
\hline
\end{tabular}
}
}

\vspace{0.25cm}

\subfloat[FRR$_{0.002\%}$ (\%)]{
\resizebox{0.7\textwidth}{!}{%
\begin{tabular}{cccccccc}
\hline
Experiment & Gaze Length & EMA & PIA & SF1 & \multicolumn{2}{c}{SF2} & EF1 \\
\cline{6-7}
 & & & & & Rank-Opt & Rank-1-Opt & \\
\hline
\multirow{4}{*}{1} 
~& 5 & 98.24 & {\multirow{3}{*}{2$^\dag$}}  & 1.93 & 2.03 & 2.03 & 0.4\\
~& 10 & 96.25 &~& 1.64 & 1.7 & 1.7 & 0.34 \\
~& 15 & 92.14 &~& 1.27 & 1.32 & 1.32 & 0.3 \\
~& 20 & 87.73 &~& 0.9 & 1.07 & 1.07 & 0.26 \\
\hline
\multirow{4}{*}{2} 
~& {\multirow{3}{*}{Same as above}} & {\multirow{3}{*}{Same as above}} & {\multirow{3}{*}{1.11$^\dag$}} & 0.9 & 1.11 & 1.11 & 0.3 \\
~& ~ &  &~& 0.77 & 0.83 & 0.83 & 0.34 \\
~& ~ & &~& 0.66 & 0.73 & 0.73 & 0.26 \\
~& ~ & &~& 0.55 & 0.61 & 0.61 & \textbf{0.22} \\
\hline
\end{tabular}
}
}

\label{tab:results}
\end{table*}

\section{Results}

We present the authentication performance under various experimental conditions with multiple fusion approaches in Table~\ref{tab:results}.
The EF2 approach is excluded from the table as it is mathematically equivalent to SF1 with a 0.5 weighting.
All results were obtained using the random saccade (jumping dot) task, using the entire training set of 6,747 users and the complete testing set of 2,455 users. 
The model was trained for 100 epochs with a mini-batch size of $m = 256$ samples, and both enrollment and verification were performed using either 5/10/20 seconds of gaze data mentioned in the ``Gaze Length" column of the tables.
In Experiment-1, a single periocular image per eye is used for each authentication attempt, whereas five images are used in Experiment-2.

PIA approach consistently outperforms EMA across both experimental setups. 
However, multimodal fusion reliably yields superior biometric performance compared to any individual modality. 
As shown in Table \ref{tab:results},  EF1, the linear transformation-based fusion approach, achieves the highest performance among all fusion strategies.


Intraclass correlation coefficient (ICC) has been established as a metric to measure temporal persistence \cite{friedman2017method}. 
Previous studies have demonstrated a strong relationship between biometric performance and ICC \cite{raju2024temporal,lohr2024baseline}. 
If the distribution of the biometric features is not normal, Kendall’s Coefficient of Concordance (KCC) \cite{field2005kendall} is a non-parametric alternative for ICC \cite{raju2024temporal}. 
In our study, a substantial number of features exhibited non-normal distributions (see supplementary materials for details), making KCC the more appropriate choice.

\begin{figure}[htbp]
\centering
\includegraphics[width=0.45\textwidth]{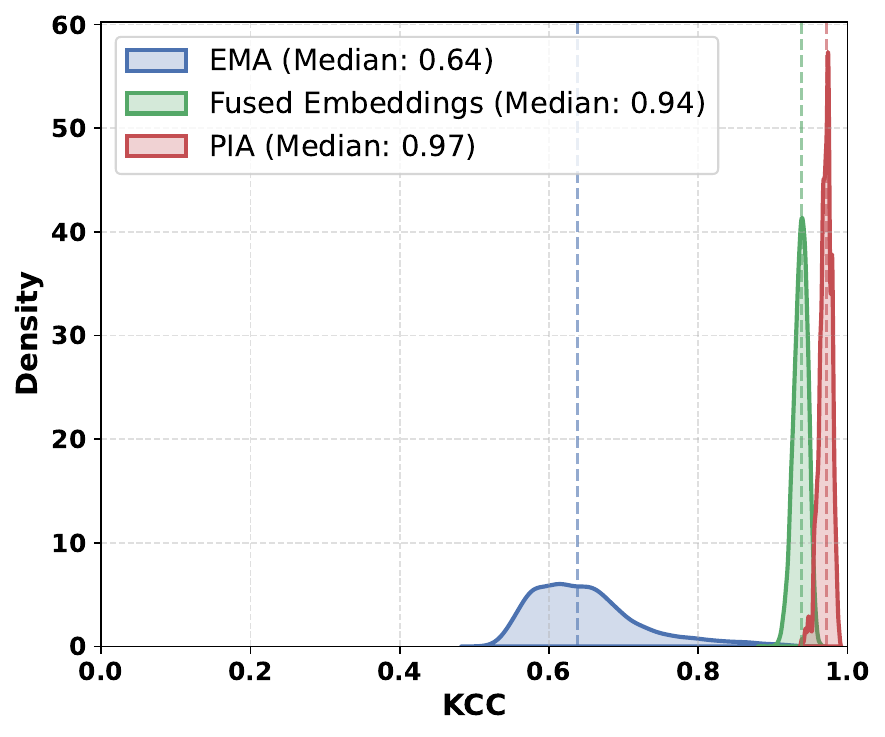}
\caption{KCC Distribution of Embeddings}
\label{fig:icc}
\end{figure}

Accordingly, we computed KCC values separately for the learned gaze embeddings, periocular embeddings, and our optimal fusion method. Figure \ref{fig:icc} presents the distribution of KCC values across these modalities. Based on prior findings, we hypothesized that the fusion-based approach (EF1), which offers the highest biometric performance, would also demonstrate the greatest temporal persistence, reflected in higher KCC values.
However, the results revealed that PIA demonstrated the highest median KCC, which was an unexpected outcome.
More details about the relation between KCC and biometric performance are added in the supplementary materials.

\section{Discussion}

\subsection{RQ Analysis}

\paragraph{(RQ1) Multimodal Authentication $>$ Individual Modality}

Multimodal fusion consistently outperforms individual modalities regarding both EER and FRR$_{0.002\%}$. 
Both the EMA and PIA exhibit significantly higher EER and FRR values across all embedding lengths, indicating comparatively lower authentication performance than the multimodal authentication system. 
In contrast, the employed fusion methods demonstrate substantial improvements in most of the cases (other than SF2 with gaze length = 5 seconds scenarios, Red highlighted in Table \ref{tab:results}), with EF1 achieving the lowest FRR$_{0.002\%}$ = 0.22\% with 20 seconds of gaze data and five-periocular images (EER = 0.05) in experiment-2.

Notably, our achieved performance met the FIDO requirements for user authentication performance. 
As specified by the FIDO biometric standard, the target is an FRR$_{0.002\%}$ of 3\% with a maximum verification duration of 30 seconds. 
In comparison, our approach achieved an FRR$_{0.002\%}$ of just 0.22\% with a verification duration of 20 seconds only.
These results highlight the effectiveness of combining gaze and periocular cues for enhanced authentication performance.

\paragraph{(RQ2) Best Fusion Approach: Fused Embeddings}
  
Among all evaluated methods, embedding-level linear transformation-based fusion EF1 consistently achieved the lowest EER \& FRR$_{0.002\%}$ across all embedding lengths.
This exceptional performance can be attributed to EF1's direct concatenation of L2-normalized gaze and periocular embeddings, followed by a linear transformation via a shallow MLP, which preserves complementary information from both modalities without introducing unnecessary complexity. 
Furthermore, the use of all training data in a single batch, combined with consistent validation on a disjoint subject set, ensured robust generalization. 
These design choices might make EF1 the most effective fusion strategy in this study.

\paragraph{(RQ3) Effect of Embedding Aggregation}

A comparative analysis between Experiment 1 and Experiment 2 reveals substantial performance improvements driven by embedding aggregation. 
In Experiment 1, the multimodal system operated under minimal input conditions from PIA by using only a single periocular image per eye, introducing high variability of the learned representations. 
In contrast, Experiment 2 employed multiple (five) periocular images per eye and averaged their embeddings, leading to more stable and representative features. 
This approach significantly lowered both EER and FRR$_{0.002\%}$, demonstrating the effectiveness of embedding aggregation in reducing variability in periocular data.
Furthermore, within each experiment, increasing the embedding length of gaze data consistently improved performance. 
Longer embedding provided richer behavioral cues and more reliable gaze dynamics, contributing to enhanced authentication performance.

\subsection{Impact of Large-Scale Dataset}
A critical factor contributing to the strong performance of our embedding-level fusion method (EF1) is the use of the large-scale GazePro dataset, comprising over 9,000 participants. 
Notably, this is a significantly larger subject pool than those found in prior ET datasets commonly used for user authentication research, such as GazeBase \cite{gazebase}, GazeBaseVR \cite{gazebasevr}, and the dataset from Makowski et al. \cite{makowski2020biometric}. 
The breadth of GazePro enables more rigorous and representative evaluations, allowing us to assess the generalization and scalability of fusion-based approaches across a wider demographic. 
This scale is especially important in authentication systems, where performance often varies with population diversity. 

\subsection{Single-device Data Acquisition}
Our study incorporates a single-device-based pipeline that captures both periocular images and eye movement signals simultaneously, which offers significant advantages for multimodal user authentication. 
It ensures temporal alignment between the two authentication modalities, enabling accurate and efficient feature fusion.
Since the same cameras are used for both periocular image capture and gaze estimation, there is no additional hardware complexity.
Additionally, this native nature facilitates continuous authentication, making it well-suited for deployment in XR devices.

\subsection{Gaze via Optical Axis Estimation}

All results reported in this study were obtained using gaze estimates based solely on the optical axis, without incorporating the visual axis estimation as did in prior research \cite{lohr2024baseline}. 
This design choice reflects a practical and calibration-free setup, aligning with the goals of real-world. 
Unlike the visual axis which typically requires a personalized calibration procedure, the optical axis can be computed directly from a generalized appearance-based ET model.
While the visual axis is often considered more accurate for point-of-gaze estimation in interactive systems, our findings demonstrate that robust user authentication is achievable using only the optical axis. 
While we currently expect the use of the visual axis to achieve even further improvement of resulting ocular authentication, this approach is beyond the scope of current work.

\subsection{Limitations \& Future Work}
When we used multiple periocular images, they were mostly taken from consecutive frames and so would be highly correlated in the absence of eye movement.
It might, therefore, be possible to achieve higher performance with PIA alone by spreading the image captures over a longer time period to reduce their correlations. 
It is important to indicate that the use of additional periocular images is more computationally expensive, which is a very important consideration for untethered XR devices.

While our study explored multiple fusion strategies, we did not incorporate more advanced techniques, such as cross-attention mechanisms or mixture-of-experts models. 
These sophisticated fusion architectures could potentially capture deeper inter-modal relationships and further improve user authentication accuracy. 
Future work may benefit from investigating such methods to push the limits of multimodal user authentication fusion.

Regarding KCC distribution, the results revealed that PIA exhibited the highest median KCC—-- an unexpected outcome based on the prior research. 
This finding indicates the need for future research to investigate why PIA demonstrates greater temporal persistence, and whether the fusion process introduces factors beyond straightforward complementarity, potentially altering or diminishing the intrinsic stability of the individual modalities.

\section{Conclusion}

In this study, we have proposed an ocular multimodal user authentication system combining two complementary modalities from a unified pipeline.
We conducted extensive experiments on a large-scale, in-house dataset comprising nearly ten thousand subjects to evaluate the system's authentication performance.
We employed the state-of-the-art EKYT architecture for EMA.
We also employed several fusion techniques, with linear transformation-based fusion being the best-performing. 
Our key finding demonstrates that the proposed multimodal authentication system consistently outperforms the unimodal approaches in most of the cases, often by a substantial margin.
The best-case authentication performance achieved an FRR$_{0.002\%}$ of 0.22\%, while the corresponding EER is the lowest overall 0.05\% in experiment setup-2, significantly better than any unimodal baseline.
Furthermore, we emphasized the importance of information sample length--- eye movement length (for EMA) and the number of periocular images (for PIA)—in influencing authentication outcomes. 
Specifically, longer enrollment and verification durations were found to yield improved user authentication accuracy.

\bibliographystyle{unsrt}
\bibliography{ms.bib}

\end{document}


\title{Supplementary Material for \textit{Ocular Authentication: Fusion of Gaze and Periocular Modalities}}

\author{
    Dillon~Lohr\textsuperscript{1}, 
    Michael J. Proulx\textsuperscript{1}, 
    Mehedi Hasan Raju\textsuperscript{2},  
    Oleg V. Komogortsev\textsuperscript{1,2}\\
    \textsuperscript{1}Meta Reality Labs Research, Redmond, WA, USA, 
    \textsuperscript{2}Texas State University, San Marcos, Texas, USA \\
    {\tt\small dlohr@meta.com, michaelproulx@meta.com, m.raju@txstate.edu, ok@txstate.edu}
}

\maketitle
\thispagestyle{empty}

\section*{Purpose of this document}
This supplementary material provides additional details about the properties of the embeddings generated in our study, which could not be included in the main manuscript due to space constraints.

\section{Properties of Embeddings}

To assess normality, we followed the normality test used by Friedman et al. \cite{friedman2021angular}, comparing the skewness and excess kurtosis of each empirical feature distribution against a population of random normal distributions.
Tables 1, 2, and 3 report the results of this analysis, showing (i) how many of the learned embeddings follow a normal distribution, (ii) intercorrelations among the embeddings, and (iii) the reliability of the embeddings for the periocular-image-based authentication (PIA), eye-movement authentication (EMA), and fused embeddings (EF1) approaches, respectively.
EF1 was performed using an MLP with no hidden layer (i.e., a linear transformation), resulting in 256-dimensional embeddings.
so, EMA learns 128-dimensional embeddings, whereas PIA and EF1 both are 256. 

The tables include the following columns: Val. Fold, Num. of Eye Images, and Gaze Data Length (seconds).

\textit{Val. Fold} indicates the fold used for validation (ranging from 0 to 9). The dataset was initially split into training and test sets; the training set was further divided into 10 folds.
\textit{Num. of Eye Images} specifies the number of left+right eye image pairs used to generate the periocular embeddings (either 1 or 5). Embeddings are computed by averaging each feature across all images, consistent with the procedure used for gaze embeddings.
\textit{Gaze Data Length (seconds)} refers to the duration of gaze data used for embedding generation (5, 10, 15, or 20 seconds). Since one embedding is derived per 5-second window, these durations yield 1 to 4 embeddings. As with periocular features, the final gaze embeddings are obtained by averaging each feature across windows.

Different combinations of values from these columns result in a variety of case analyses, enabling a comprehensive evaluation of the learned embeddings under varying validation folds, image counts, and gaze durations.

According to the tables, the median intercorrelation is notably low, while the median KCC remains consistently high for both PIA and EF1. This supports the claim made by \cite{raju2024temporal} that embeddings that are weakly correlated with higher measurement reliability are crucial for biometric performance.

\section{Relation between KCC and EER}
\begin{figure}[htbp]
\centering
\includegraphics[width=0.45\textwidth]{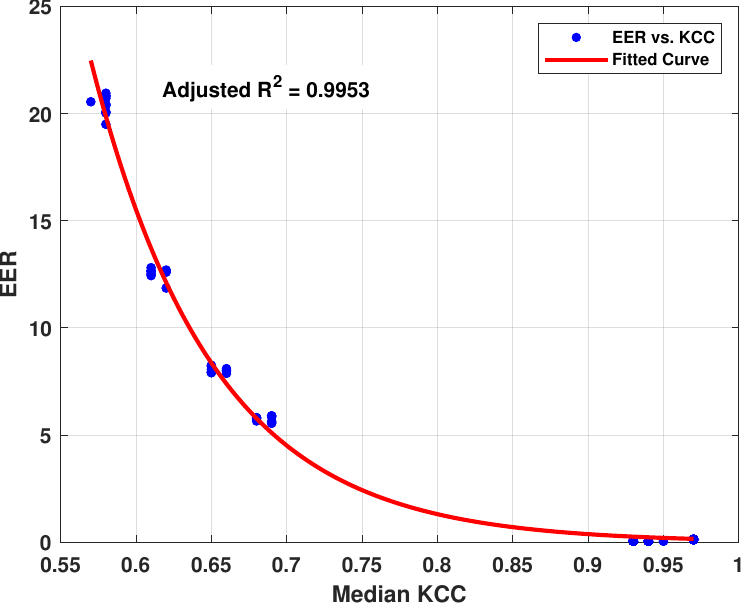}
\caption{Observations of EER vs KCC. The Figure takes the median KCC across the n-embedding, where n=128 for EMA, 256 for PIA and Fused embeddings. An exponential curve is fit to the observations to highlight the trend. Adjusted $R^2$ is annotated in the figure.}
\label{fig:eervskcc}
\end{figure}

In Figure \ref{fig:eervskcc}, we demonstrated the relation between the temporal persistence measured by KCC and the biometric performance. Similar to prior research \cite{friedman2017method,lohr2024baseline}, a strong relation is seen in our observation as well. Here we have plotted all KCC along with their EER value from all three cases: EMA, PIA, and EF1.
There is a strong relation between KCC and EER, validated by near near-perfect adjusted $R^2$ value.

%

\begin{longtable}{cccccccc}
\caption{Assessment of normality, intercorrelations, and reliability of the 256-dimensional learned embeddings from the periocular image-based authentication approach. Minimum intercorrelation across all experiments was 0.00 and is therefore omitted from the table.}\\
\hline
\multirow{2}{*}{Num. of Eye Images} & \multirow{2}{*}{Normal} & \multicolumn{2}{c}{Intercorrelation} & \multicolumn{3}{c}{KCC}\\ \cline{3-7}
~&~& Median & Max & Min & Med  & Max  \\ \hline
1 & 37 & 0.14 & 0.76 & 0.94 & 0.97 & 0.98 \\ 
5 & 41 & 0.14 & 0.76 & 0.95  & 0.97 & 0.99 \\ \hline
\end{longtable}

\begin{longtable}{cccccccc}
\caption{Same as Table 1, but for the 128-dimensional embeddings learned using the eye-movement-based authentication approach.}\\
\hline
\multirow{2}{*}{Val. Fold} & \multirow{2}{*}{Gaze Length (seconds)} & \multirow{2}{*}{Normal} & \multicolumn{2}{c}{Intercorrelation} & \multicolumn{3}{c}{KCC}\\ \cline{4-8} 
   &  &   & Median & Max & Min & Med  & Max  \\ \hline
\multirow{4}{*}{0}
    &   5   &   84  &   0.04    &   0.49   &    0.52    &   0.58    &   0.82 \\
    &   10  &   92  &   0.05    &   0.55   &    0.54    &   0.61    &   0.88 \\
    &   15  &   90  &   0.05    &   0.59   &    0.57    &   0.65    &   0.9  \\
    &   20  &   84  &   0.06    &   0.62   &    0.61    &   0.68    &   0.91 \\ \hline
   
\multirow{4}{*}{1}
    &    5  &   95  &   0.04    &   0.45    &   0.53    &   0.58    &   0.84 \\
    &   10  &   95  &   0.05    &   0.46    &   0.57    &   0.61    &   0.88 \\
    &   15  &   89  &   0.05    &   0.47    &   0.58    &   0.65    &   0.9 \\ 
    &   20  &   85  &   0.06    &   0.46    &   0.61    &   0.68    &   0.92 \\ \hline
   
\multirow{4}{*}{2}
    &   5   &   98  &   0.04    &   0.46    &   0.53    &   0.58    &   0.81 \\  
    &   10  &   93  &   0.05    &   0.49    &   0.57    &   0.61    &   0.87 \\  
    &   15  &   85  &   0.06    &   0.52    &   0.59    &   0.65    &   0.9 \\  
    &   20  &   80  &   0.06    &   0.54    &   0.61    &   0.69    &   0.91 \\ \hline
   
\multirow{4}{*}{3}
    &   5   &   95  &   0.04    &   0.46   &    0.52    &   0.58    &   0.87 \\
    &   10  &   96  &   0.05    &   0.52   &    0.54    &   0.62    &   0.91 \\  
    &   15  &   93  &   0.05    &   0.56   &    0.59    &   0.66    &   0.93 \\  
    &   20  &   84  &   0.06    &   0.58   &    0.61    &   0.69    &   0.93 \\ \hline
   
\multirow{4}{*}{4}
&   5   &   94  & {0.04}    & 0.48   & {0.53} & {0.58} & 0.86 \\  
&   10& 97& {0.05}    & 0.57   & {0.55} & {0.62} & 0.9 \\  
   & 15& 91& {0.06}    & 0.6    & {0.58} & {0.66} & 0.92 \\  
   & 20& 87& {0.06}    & 0.63   & {0.62} & {0.69} & 0.93 \\ \hline

\multirow{4}{*}{5}
& 5& 92& {0.04}    & 0.45   & {0.54} & {0.58} & 0.85 \\  
   & 10& 101& {0.05}    & 0.49   & {0.56} & {0.61} & 0.89 \\  
   & 15& 88& {0.05}    & 0.5    & {0.6}  & {0.65} & 0.91 \\  
   & 20& 82& {0.06}    & 0.5    & {0.62} & {0.68} & 0.93 \\ \hline

\multirow{4}{*}{6}
& 5& 93& {0.05}    & 0.53   & {0.53} & {0.58} & 0.8 \\  
   & 10& 95& {0.05}    & 0.55   & {0.55} & {0.62} & 0.86 \\  
   & 15& 85& {0.06}    & 0.55   & {0.6}  & {0.66} & 0.89 \\  
   & 20& 82& {0.07}    & 0.56   & {0.62} & {0.69}  & 0.91 \\ \hline
   
\multirow{4}{*}{7}
& 5& 98& {0.04}    & 0.57   & {0.53} & {0.57} & 0.82 \\  
   & 10& 95& {0.05}    & 0.6    & {0.56} & {0.61} & 0.87 \\  
   & 15& 97& {0.05}    & 0.6    & {0.59} & {0.65}  & 0.89 \\  
   & 20& 81& {0.06}    & 0.59   & {0.61} & {0.68} & 0.91 \\ \hline
   
\multirow{4}{*}{8}
& 5& 92& {0.05}    & 0.41   & {0.54} & {0.58} & 0.83 \\  
   & 10& 102& {0.05}    & 0.46   & {0.55} & {0.62} & 0.88 \\  
   & 15& 91& {0.06}    & 0.51   & {0.59}  & {0.66} & 0.9  \\  
   & 20& 90& {0.07}    & 0.55   & {0.61} & {0.69} & 0.92 \\ \hline
   
\multirow{4}{*}{9}
& 5& 91& {0.04}    & 0.45   & {0.52} & {0.58} & 0.84  \\  
   & 10& 101& {0.05}    & 0.51   & {0.55}  & {0.61} & 0.89 \\  
   & 15& 92& {0.06}    & 0.55   & {0.58} & {0.65} & 0.92 \\  
   & 20& 89& {0.06}    & 0.6    & {0.6} & {0.69} & 0.93 \\ \hline

\end{longtable}

\begin{longtable}{cccccccccc}
\caption{Same as Tables 1 and 2, but for 256-dimensional Fused embeddings obtained by fusing periocular and eye-movement-based embeddings.}\\
\hline
\multirow{2}{*}{Val. Fold} & \multirow{2}{*}{Num. of Eye Images} & \multirow{2}{*}{Gaze Length (seconds)} & \multirow{2}{*}{Normal} & \multicolumn{2}{c}{Intercorrelation} & \multicolumn{3}{c}{KCC}  \\ \cline{5-9} 
   &  &  &   & {Median}   & Max   & {Min}  & {Median} & Max  \\ \hline
   
\multirow{8}{*}{0} 
  & 1 & 5  & 161 & 0.08 & 0.49 & 0.91 & 0.93 & 0.96 \\
  & 1 & 10 & 163 & 0.08 & 0.45 & 0.91 & 0.93 & 0.96 \\
  & 1 & 15 & 154 & 0.08 & 0.53 & 0.91 & 0.93 & 0.95 \\
  & 1 & 20 & 155 & 0.08 & 0.52 & 0.91 & 0.93 & 0.95 \\ \cline{2-9}
  & 5 & 5  & 161 & 0.08 & 0.49 & 0.93 & 0.95 & 0.96 \\
  & 5 & 10 & 167 & 0.07 & 0.50 & 0.93 & 0.94 & 0.96 \\
  & 5 & 15 & 149 & 0.08 & 0.61 & 0.92 & 0.94 & 0.96 \\
  & 5 & 20 & 164 & 0.08 & 0.56 & 0.92 & 0.94 & 0.96 \\ \hline

\multirow{8}{*}{1} 
  & 1 & 5  & 153 & 0.08 & 0.51 & 0.90 & 0.93 & 0.96 \\
  & 1 & 10 & 155 & 0.08 & 0.48 & 0.91 & 0.93 & 0.95 \\
  & 1 & 15 & 153 & 0.08 & 0.53 & 0.91 & 0.93 & 0.95 \\
  & 1 & 20 & 147 & 0.08 & 0.58 & 0.91 & 0.93 & 0.95 \\ \cline{2-9}
  & 5 & 5  & 152 & 0.07 & 0.49 & 0.93 & 0.94 & 0.96 \\
  & 5 & 10 & 156 & 0.08 & 0.49 & 0.93 & 0.94 & 0.96 \\
  & 5 & 15 & 159 & 0.07 & 0.52 & 0.92 & 0.94 & 0.96 \\
  & 5 & 20 & 153 & 0.08 & 0.56 & 0.92 & 0.94 & 0.96 \\ \hline

\multirow{8}{*}{2} 
  & 1 &  5 & 161 & 0.08 & 0.48 & 0.90 & 0.93 & 0.96 \\  
  & 1 & 10 & 156 & 0.08 & 0.47 & 0.91 & 0.93 & 0.96 \\  
  & 1 & 15 & 151 & 0.08 & 0.46 & 0.91 & 0.93 & 0.95 \\  
  & 1 & 20 & 152 & 0.08 & 0.45 & 0.90 & 0.93 & 0.95 \\ \cline{2-9}
  & 5 &  5 & 163 & 0.08 & 0.47 & 0.92 & 0.94 & 0.96 \\  
  & 5 & 10 & 157 & 0.08 & 0.48 & 0.92 & 0.94 & 0.96 \\  
  & 5 & 15 & 147 & 0.08 & 0.51 & 0.92 & 0.94 & 0.96 \\  
  & 5 & 20 & 160 & 0.08 & 0.50 & 0.92 & 0.94 & 0.96 \\ \hline

\multirow{8}{*}{3} 
  & 1 &  5 & 159 & 0.08 & 0.50 & 0.91 & 0.94 & 0.96 \\
  & 1 & 10 & 149 & 0.08 & 0.49 & 0.91 & 0.93 & 0.96 \\
  & 1 & 15 & 151 & 0.08 & 0.49 & 0.91 & 0.93 & 0.95 \\
  & 1 & 20 & 142 & 0.08 & 0.49 & 0.91 & 0.93 & 0.95 \\ \cline{2-9}
  & 5 &  5 & 159 & 0.08 & 0.51 & 0.92 & 0.94 & 0.96 \\
  & 5 & 10 & 149 & 0.08 & 0.49 & 0.92 & 0.94 & 0.96 \\
  & 5 & 15 & 145 & 0.08 & 0.49 & 0.92 & 0.94 & 0.96 \\
  & 5 & 20 & 140 & 0.08 & 0.44 & 0.92 & 0.94 & 0.96 \\ \hline

\multirow{8}{*}{4} 
  & 1 & 5  & 161 & 0.08 & 0.48 & 0.91 & 0.93 & 0.96 \\
  & 1 & 10 & 154 & 0.08 & 0.46 & 0.91 & 0.93 & 0.95 \\
  & 1 & 15 & 156 & 0.08 & 0.48 & 0.91 & 0.93 & 0.95 \\
  & 1 & 20 & 166 & 0.08 & 0.47 & 0.90 & 0.93 & 0.95 \\ \cline{2-9}
  & 5 & 5  & 166 & 0.08 & 0.51 & 0.93 & 0.94 & 0.96 \\
  & 5 & 10 & 171 & 0.08 & 0.46 & 0.93 & 0.94 & 0.96 \\
  & 5 & 15 & 167 & 0.08 & 0.47 & 0.92 & 0.94 & 0.96 \\
  & 5 & 20 & 162 & 0.08 & 0.45 & 0.92 & 0.94 & 0.96 \\ \hline

\multirow{8}{*}{5} 
  & 1 & 5  & 146 & 0.08 & 0.47 & 0.91 & 0.93 & 0.96 \\
  & 1 & 10 & 157 & 0.08 & 0.45 & 0.91 & 0.93 & 0.95 \\
  & 1 & 15 & 147 & 0.08 & 0.50 & 0.91 & 0.93 & 0.95 \\
  & 1 & 20 & 149 & 0.08 & 0.56 & 0.90 & 0.93 & 0.95 \\ \cline{2-9}
  & 5 & 5  & 167 & 0.07 & 0.46 & 0.92 & 0.95 & 0.96 \\
  & 5 & 10 & 151 & 0.08 & 0.45 & 0.92 & 0.94 & 0.96 \\
  & 5 & 15 & 152 & 0.08 & 0.50 & 0.92 & 0.94 & 0.96 \\
  & 5 & 20 & 150 & 0.08 & 0.54 & 0.92 & 0.94 & 0.96 \\ \hline

\multirow{8}{*}{6}& 1& 5& 160& 0.08 & 0.52 & 0.91 & 0.93 & 0.96 \\  
   & 1& 10& 156& 0.08 & 0.45 & 0.91 & 0.93 & 0.95 \\  
   & 1& 15& 164& 0.08 & 0.47 & 0.90 & 0.93 & 0.95 \\  
   & 1& 20& 156& 0.08 & 0.45 & 0.90 & 0.93 & 0.95 \\  \cline{2-9}
   & 5& 5& 164& 0.07 & 0.48 & 0.92 & 0.94 & 0.96 \\  
   & 5& 10& 170& 0.08 & 0.45 & 0.92 & 0.94 & 0.96 \\  
   & 5& 15& 162& 0.08 & 0.48 & 0.92 & 0.94 & 0.96 \\  
   & 5& 20& 149& 0.08 & 0.50 & 0.91 & 0.94 & 0.96 \\ \hline

\multirow{8}{*}{7}& 1& 5& 157& 0.08 & 0.49 & 0.91 & 0.93 & 0.96 \\  
   & 1& 10& 147& 0.08 & 0.55 & 0.91 & 0.93 & 0.96 \\  
   & 1& 15& 140& 0.08 & 0.52 & 0.91 & 0.93 & 0.95 \\  
   & 1& 20& 146& 0.08 & 0.49 & 0.91 & 0.93 & 0.95 \\  \cline{2-9}
   & 5& 5& 155& 0.08 & 0.48 & 0.92 & 0.94 & 0.97 \\  
   & 5& 10& 152& 0.08 & 0.50 & 0.92 & 0.94 & 0.96 \\  
   & 5& 15& 156& 0.08 & 0.53 & 0.92 & 0.94 & 0.96 \\  
   & 5& 20& 146& 0.08 & 0.50 & 0.92 & 0.94 & 0.96 \\ \hline

\multirow{8}{*}{8}& 1& 5& 152& 0.08 & 0.51 & 0.91 & 0.93 & 0.95 \\  
   & 1& 10& 161& 0.08 & 0.49 & 0.91 & 0.93 & 0.95 \\  
   & 1& 15& 151& 0.08 & 0.49 & 0.90 & 0.93 & 0.95 \\  
   & 1& 20& 156& 0.08 & 0.48 & 0.88 & 0.93 & 0.95 \\  \cline{2-9}
   & 5& 5& 162& 0.08 & 0.51 & 0.93 & 0.94 & 0.96 \\  
   & 5& 10& 165& 0.08 & 0.50 & 0.92 & 0.94 & 0.96 \\  
   & 5& 15& 158& 0.08 & 0.50 & 0.92 & 0.94 & 0.96 \\  
   & 5& 20& 159& 0.08 & 0.51 & 0.92 & 0.94 & 0.96 \\ \hline

\multirow{8}{*}{9}& 1& 5& 156& 0.08 & 0.54 & 0.91 & 0.94 & 0.96 \\  
   & 1& 10& 154& 0.08 & 0.46 & 0.89 & 0.93 & 0.96 \\  
   & 1& 15& 161& 0.08 & 0.46 & 0.90 & 0.93 & 0.95 \\  
   & 1& 20& 148& 0.08 & 0.46 & 0.90 & 0.93 & 0.95 \\  \cline{2-9}
   & 5& 5& 155& 0.07 & 0.55 & 0.92 & 0.95 & 0.97 \\  
   & 5& 10& 151& 0.07 & 0.53 & 0.91 & 0.94 & 0.96 \\  
   & 5& 15& 149& 0.08 & 0.49 & 0.92 & 0.94 & 0.96 \\  
   & 5& 20& 158& 0.08 & 0.47 & 0.92 & 0.94 & 0.96 \\ \hline

\end{longtable}

\bibliographystyle{unsrt}
\bibliography{supp.bib}